\documentclass{article} 
\usepackage{iclr2022_conference,times}

\usepackage{amsmath,amsfonts,bm}

\def\eqref#1{equation~\ref{#1}}

\def\1{\bm{1}}

\DeclareMathAlphabet{\mathsfit}{\encodingdefault}{\sfdefault}{m}{sl}
\SetMathAlphabet{\mathsfit}{bold}{\encodingdefault}{\sfdefault}{bx}{n}

\usepackage{hyperref}
\usepackage{url}
\usepackage{multirow, makecell}
\usepackage{graphicx}
\usepackage{wrapfig}
\usepackage{spverbatim}
\usepackage{caption}
\usepackage{subcaption}
\usepackage{fancyvrb}
\usepackage{listings}

\title{Understanding HTML with Large Language Models}

\author{%
Izzeddin Gur, Ofir Nachum, Yingjie Miao, Mustafa Safdari, Austin Huang \\ 
\textbf{Aakanksha Chowdhery, Sharan Narang, Noah Fiedel, Aleksandra Faust}\\
Google Research\\
\texttt{\{\href{mailto:izzeddin@google.com}{izzeddin},ofirnachum,yingjiemiao,msafdari,austinvhuang} \\
\texttt{chowdhery,sharannarang,nfiedel,sandrafaust\}@google.com} \\
}

\newcommand*{\llm}{LLM}
\newcommand*{\llms}{LLMs}

\newcommand*{\htmlpage}{HTML}

\newcommand*{\taskClassification}{\textit{Semantic Classification}}
\newcommand*{\taskDescription}{\textit{Description Generation}}
\newcommand*{\taskNavigation}{\textit{Autonomous Web Navigation}}
\newcommand*{\modelC}{WebC-}
\newcommand*{\modelD}{WebD-}
\newcommand*{\modelN}{WebN-}

\iclrfinalcopy
\begin{document}

\maketitle

\begin{abstract}
Large language models (LLMs) have shown exceptional performance on a variety of natural language tasks.
Yet, their capabilities for HTML understanding -- i.e., parsing the raw HTML of a webpage, with applications to automation of web-based tasks, crawling, and browser-assisted retrieval -- 
have not been fully explored.
We contribute HTML understanding models (fine-tuned \llms) and an in-depth analysis of their capabilities under three tasks: (i) \taskClassification\ of HTML elements, (ii) \taskDescription\ for HTML inputs, and (iii) \taskNavigation\ of HTML pages.
While previous work has developed dedicated architectures and training procedures for HTML understanding, we show that \llms\ pretrained on standard natural language corpora transfer remarkably well to HTML understanding tasks. 
For instance, fine-tuned \llms\ are 12\% more accurate at semantic classification compared to models trained exclusively on the task dataset.
Moreover, when fine-tuned on data from the MiniWoB benchmark, \llms\ successfully complete 50\% more tasks using 192x less data compared to the previous best supervised model.
Out of the LLMs we evaluate, we show evidence that T5-based models are ideal due to their bidirectional encoder-decoder architecture.
To promote further research on \llms\ for HTML understanding, we create and open-source a large-scale HTML dataset distilled and auto-labeled from CommonCrawl.\footnote{See visualizations of the results at \url{https://sites.google.com/view/llm4html/home}.}
\end{abstract}

\section{Introduction}
\label{lab:intro}
Web crawling~\citep{olston2010web}, form-filling~\citep{diaz2013user,gur2021environment}, or information retrieving web agents~\citep{nogueira2016end} are important for both automating and assisting users in web-based tasks. These and similar applications rely on models that can search for specific content or controls on a web page as well as navigate a website autonomously. Since a web page in its raw form is represented as an HTML-based text sequence, the success of models for web-based tasks relies on their ability to understand HTML semantics, structure, and embedded interactions.

The predominant approach to web automation and HTML understanding is to train specialized models, i.e., gathering application-specific datasets and designing neural network (NN) architectures to leverage inductive biases of the HTML's structure; see, e.g.,~\citet{liu2018reinforcement,toyama2021androidenv,gur2021environment,humphreys2022data}. 
However, both dataset collection and neural architecture design are expensive, time-consuming, and require highly-specialized, domain-specific knowledge.

Meanwhile, in the natural language processing (NLP) literature, large language models (\llms) have emerged as a solution to the difficulties of dataset collection and specialized NN design~\citep{kaplan2020scaling,bommasani2021opportunities}. 
A popular paradigm in NLP is to take an off-the-shelf LLM -- pretrained on a large text corpus via an unsupervised and task-agnostic learning objective -- and either fine-tune or prompt the LLM on a small task-specific dataset. 
This paradigm has shown exceptional performance on a variety of NLP tasks~\citep{xue2020mt5,brown2020language,austin2021program}.
Whether LLMs can be applied to HTML understanding -- especially given the much larger context and sequence lengths -- remains an under-explored question.

\begin{figure}
     \centering
     \begin{subfigure}[b]{0.35\textwidth}
         \centering
         \includegraphics[width=\textwidth]{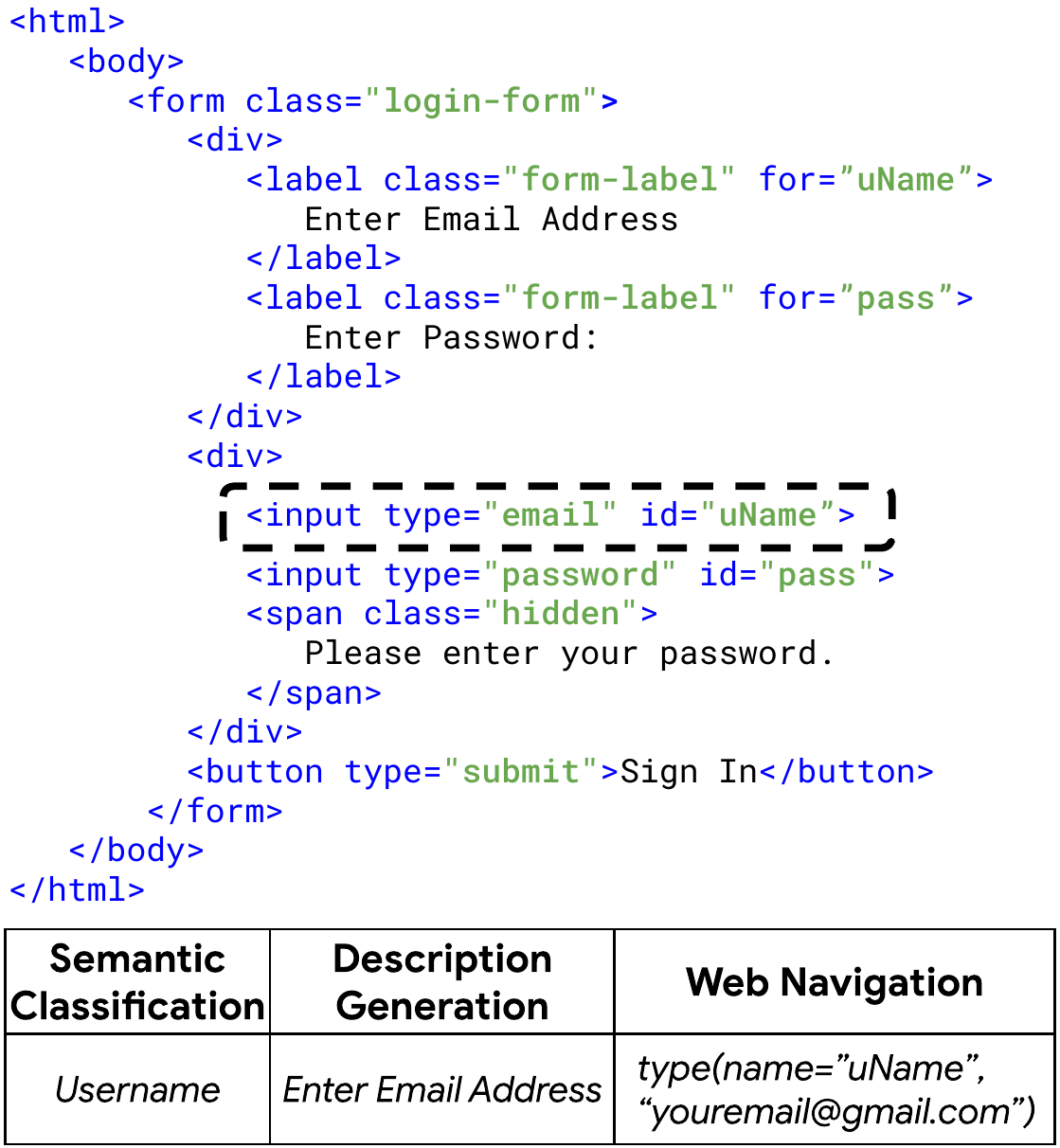}
         \caption{\small} 
         \label{diagram:problem}
     \end{subfigure}
     \hfill 
     \begin{subfigure}[b]{0.6\textwidth}
         \centering
         \includegraphics[width=\textwidth]{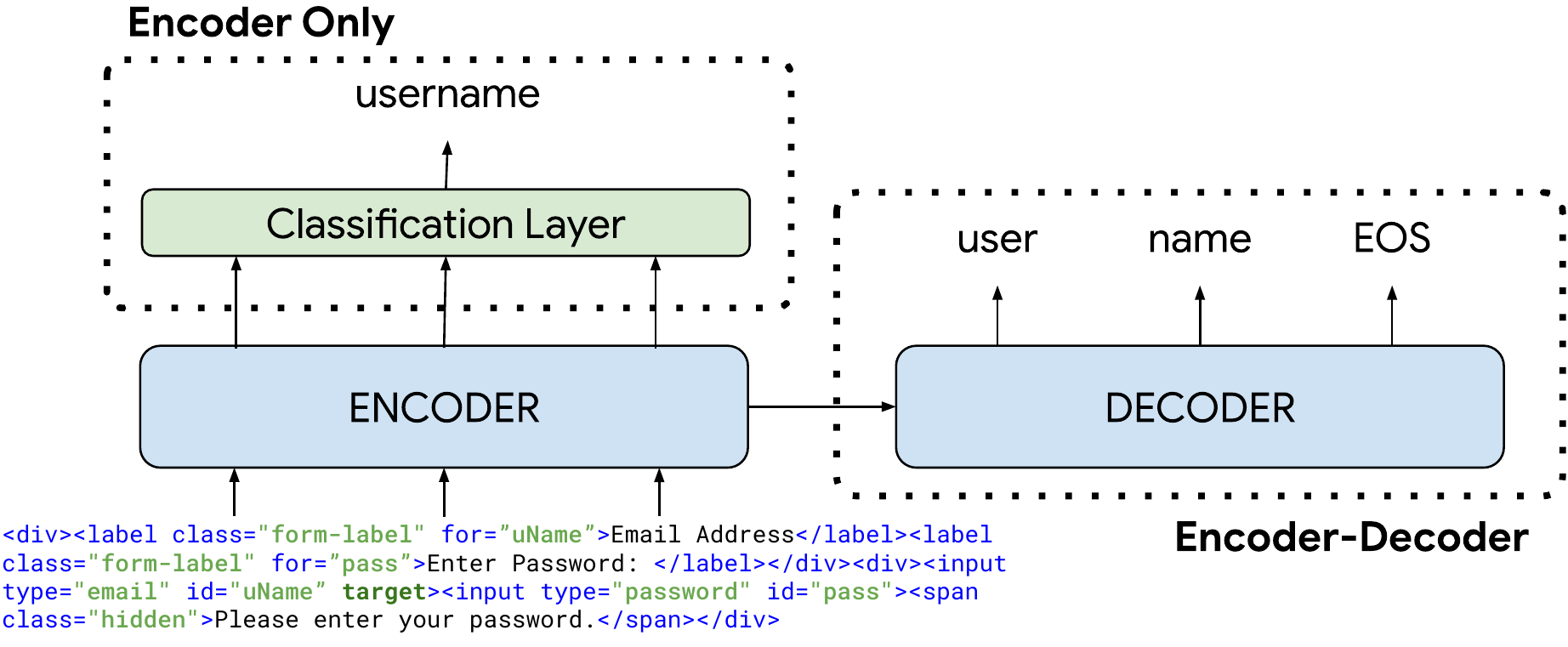}
         \caption{\small} 
         \label{diagram:model}
     \end{subfigure}
    
    \caption{\small a) HTML example page with a highlighted salient element, an element of interest (dashed box). All canonical tasks evaluate a distinct interaction with this element, either by classifying it as one of a set of categories, generating a text description of its purpose, or applying an action as part of a sequential navigation of a multi-page website. b) \llm\ architectures overview. Dashed boxes denote sub-modules that are specific to either encoder-only or encoder-decoder models. For encoder-only models, we add an extra classification layer. Decoder-only models (not in the diagram) are similar to encoder-decoder models, the main difference is that the HTML snippet is fed to the decoder and processed from left-to-right.}
    \label{fig:1}
\end{figure}

In this paper, we investigate whether LLMs can be applied to HTML understanding to produce better-performing, more sample-efficient HTML understanding models and without the need for custom NN architecture design. 
To that end, we present a suite of three benchmarking tasks for HTML understanding that capture the essence of these applications and require understanding both structure and content. First, we devise \taskClassification\ as a task that requires a model to classify a given HTML element into one of a set of categories, such as address, email, password etc., with application to automated form-filling. 
Second, we present \taskDescription, a label-extraction task where a model is given an HTML snippet and is asked to produce a natural language description. For instance for an email field, the description might be ``Please enter your email address.'' Note that in the majority of web pages, this connection between input elements and description content is only implicit in the raw HTML code and inferring such links is a prerequisite for higher-level navigation objectives.
The third task is \taskNavigation\ \citep{shi2017world}. A model is presented with an HTML page paired with a natural language command and must apply appropriate actions on a sequence of HTML pages to satisfy the command. See Figure \ref{diagram:problem} for a simplified example of these tasks.

With these benchmark tasks in hand, we evaluate the transfer capabilities of a variety of pretrained LLMs (Table~\ref{table:paper_summary}), varying in architecture (encoder-only, encoder-decoder, or decoder-only), model size (from 24.6M to 62B parameters), and training data corpora (both including and excluding pretraining NLP and HTML corpus). While prior work universally pre-parses the HTML as input to the model \citep{gur2021environment, liu2018reinforcement,nakano2021webgpt}, ours -- to the best of our knowledge -- is the first work that uses raw, unprocessed HTML. 
Our results show that LLMs demonstrate a remarkable level of HTML understanding across all tasks, with up to $192\times$ more sample-efficiency than models trained from scratch, and achieving a new SoTA for supervised learning on the MiniWoB benchmark suite~\citep{shi2017world}. The encoder-decoder architectures with bi-directional attention show the best performance across the board even when their pretraining does not include HTML. In addition, we show that the performance scales sub-linearly with the model size.

The broader objective of this research is to advance the integration of LLMs with autonomous web agents. It has only been in the last year that researchers have begun to utilize LLMs outside of NLP and integrate them as core capabilities in autonomy (\cite{lu2021pretrained, ahn2022can}). In this context, LLMs are reasoning engines for sequential decision making agents interacting with environments.

The present work is the first in the research literature to embed an LLM and train it as an agent for autonomous web navigation. This requires new implementations to adapt LLM training for behavior cloning in addition to designing interfaces for integrating text generation into a perception-compute-action cycle operating in a stateful web environment.  Our implementation allows us to answer new questions regarding trade-offs among various model characteristics.

We believe these contributions expand the scope of language models and connect their unique capabilities with autonomous agents for the web.
We provide a new perspective on machine learning for HTML understanding and web automation, showing that pretrained LLMs can achieve significant performance on such tasks, reducing the need for specialized architectures and training protocols.
To encourage further research in this direction, we open sourced \footnote{\url{https://console.cloud.google.com/storage/browser/gresearch/webllm}} model weights for agents used in the WoB environment and our dataset for description generation.

\section{Related Work}
\label{sec:related_work}

\textbf{HTML Understanding}
Autonomous web navigation has been a popular application for neural network models, and a variety of works propose simulated websites for training web-based agents, with application to task fulfillment \citep{yao2022webshop, gur2021environment, burns2022interactive, mazumder2020flin, shi2017world, liu2018reinforcement} as well as information retrieval or question-answering \citep{adolphs2021boosting, nogueira2016end}. 
Simulated websites provide an easy way to evaluate models online, and for this reason we use the existing MiniWoB benchmark~\citep{shi2017world} for our web navigation setting. 
However, it is still important to have a mechanism for evaluating models on a wide variety of real-world websites. This was the key motivation for generating our own dataset for the description generation task, which is distilled and auto-labeled from CommonCrawl and is a key contribution of our paper.

Alongside these benchmarks, many works have developed models for web navigation and related subtasks~\citep{pasupat2018mapping,bommasani2021opportunities,he2021actionbert,gur2021environment,humphreys2022data,liu2018reinforcement,jia2018domqnet}.
These works often rely on specialized neural network architectures that leverage inductive biases of HTML structure, or on preprocessing of HTML to make it easier to input to a model (\cite{li2021structurallm, li2021markuplm}). 
In contrast, our work takes a minimalist approach, providing HTML in text form with minimal processing and using widely-adopted transformer networks.

\textbf{LLMs and HTML}
Works that explore the intersection of LLMs and HTML generally fall into two categories. The first category uses LLMs to assist web navigation~\citep{nakano2021webgpt,yao2022webshop}, and typically relies on a custom preprocessing to map the context and structure of a web page to natural language, thus severely restricting what HTML pages the model can parse. 
The second category pretrains LLMs on a large corpora of HTML text~\citep{aghajanyan2021htlm}. However, these works typically restrict the model evaluation to standard NLP tasks, e.g., summarization and question/answering as opposed to tasks more relevant to HTML understanding and web automation.
Our work can be thought of as the reverse: We keep the pretraining of LLMs unchanged and focus on the mechanisms for transferring the pretrained LLMs to HTML-relevant tasks.

\section{Brief Background on HTML as Semi-Structured Text Data}
\label{sec:htmlbackground}
HTML is a markup language, used to organize web page \textbf{structure} and \textbf{content}.
Consider the example HTML page in Figure \ref{diagram:problem}.
This web page includes two adjacent \texttt{input} elements, one for e-mail and another for password, with their corresponding \texttt{label}s on a separate branch of the page.
These \texttt{input}s and \texttt{label}s are one of many possible \textit{elements} that serve as HTML building blocks.
Each element has a set of attributes -- key and value pair -- that describe the element's content, such as style and human-readable text.
When rendered in a browser, these attributes will be responsible for how the element is shown and where it is positioned.
In the example in Figure \ref{diagram:problem}, the first \texttt{input} has three attributes, \texttt{tag="input"}, \texttt{type="email"}, and \texttt{id="uName"}, that identify the element as an email input with an identifier (``uName'') that can be accessed programmatically. 

\section{Canonical Tasks for HTML Understanding}
\label{sec:htmlunderstanding}

We devise three canonical tasks to study HTML understanding capabilities of LLM-based web agents. These tasks require correctly interpreting both structure and content to varying degrees to make predictions, with autonomous navigation being the most challenging capability of the three.

\textbf{\taskNavigation}. This task evaluates how well a model navigates multi-page websites as a sequential decision-making problem~\citep{shi2017world,liu2018reinforcement}. 
At the beginning of an episode, the agent is given a natural language instruction, e.g. \textit{Enter the username ``lyda'' and the password ``N22t'' into the text fields and press login}.
The agent applies actions to a sequence of HTML pages, where each action is of the form \texttt{function(selector, text)}.
The \texttt{function} is one of \textit{click} or \textit{type}, \texttt{selector} is an integer pointer that uniquely identifies an element, and \texttt{text} is a text to input if the \textit{type} functionality is activated.
An episode terminates when either the page reaches a terminal state (e.g., the `sign in' button is clicked) or the maximum number of steps is reached.

\textbf{\taskClassification.} 
Many HTML understanding applications require a model that can classify HTML elements into standardized categories. For example, in automated form-filling~\citep{diaz2013user,gur2021environment}, it is useful to identify a `submit button' across many websites (e.g., shopping, flight booking, utility application) with various button representations (e.g., position, color, or text).
Thus, we formulate \taskClassification\ as classifying elements into \emph{role} categories. 
Take the example \htmlpage\ in Figure~\ref{diagram:problem} which includes two \texttt{input} elements and a submit \texttt{button}.
Let's pick the first \texttt{input} as an element of interest to be classified by the system, also called a \textit{salient element}.
The system should classify this element as \textit{username}, since it appears on a login page and it has a \texttt{label} with \textit{Email Address} which is typically associated with the username in form-filling applications.
To solve this, the system can aggregate information from multiple sources in the page -- the label that says \textit{Enter Email Address}, the \texttt{input} attributes (\textit{type=``email''} and \textit{id=``uName''}), or even the ordering of other elements in the page such as `password' and `sign in'.

\begin{table}[]
\centering
\resizebox{\textwidth}{!}{
\scriptsize
\begin{tabular}{ ccccccc  }

 &&&\multicolumn{3}{c}{\textbf{Model}}& \\ 
 \cline{4-6}
 \textbf{Task}& \textbf{Dataset} &  \textbf{Size} & \textbf{Input} & \textbf{Architecture} & \textbf{Output} & \textbf{Task Output}\\
 \hline

\multirow{2}{*}{\taskNavigation} & \multirow{2}{*}{MiniWoB Demos \citep{shi2017world}}  & \multirow{2}{*}{12K} & \multirow{2}{*}{Page} & Enc-Dec & \multirow{2}{*}{Text} & \multirow{2}{*}{Dictionary} \\
&&&&Dec&& \\\hline

 \multirow{2}{*}{\taskClassification} & \multirow{2}{*}{Annotated Shopping Webpages \citep{gur2021environment}} & \multirow{2}{*}{28K} & \multirow{2}{*}{Snippet} & \multirow{2}{*}{All} & \multirow{2}{*}{Text} & \multirow{2}{*}{Category} \\
&&&&&& \\ \hline
 
 \multirow{2}{*}{\taskDescription} & \multirow{2}{*}{CommonCrawl (new)} &  \multirow{2}{*}{85K} & \multirow{2}{*}{Snippet} & Enc-Dec & \multirow{2}{*}{Text} & \multirow{2}{*}{Text} \\
&&&&Dec& &\\ \hline
\end{tabular}}
\caption{\small Task, dataset, and model summary. All models receive raw \htmlpage. \taskNavigation\ receives the entire \htmlpage, while the other tasks receive HTML snippets extracted given salient element.}
\label{table:paper_summary}
\end{table}

\textbf{\taskDescription.} Motivated by applications in accessibility-minded web browser control~\citep{jorgensen2005web}, we formulate description generation as an extractive problem where the goal is to locate the textual description of an element in the \htmlpage\ and generate it as output.
For instance, the description of the salient element in Figure~\ref{diagram:problem} is \textit{Enter Email Address}; when rendered, this \texttt{label} will appear above the `email' \texttt{input} field.
HTML provides a large amount of flexibility, and so in general a descriptive text that appears alongside a specific element when rendered can be very far from that element when looking at the HTML plaintext. 
Thus, this task evaluates a model's ability to understand the structure of HTML as it would appear to a user, despite not having access to the rendered web page directly.

\section{Datasets}
Each of our canonical tasks requires a separate dataset, with the description generation task using a newly contributed, auto-labelled dataset based on CommonCrawl.

\textbf{\taskNavigation.} 
\label{dataset:webnav} We use the 12K demonstrations included in the publicly available MiniWoB benchmark~\citep{shi2017world}, which encompass 62 website applications ranging from email forwarding to social media interactions.
Each demonstration is a sequence of \textbf{(instruction, \htmlpage, action)} tuples.
Every element in a MiniWoB demonstration is accompanied by a reference number unique within its respective pages.
This number can be used as an element selector, making the action space unified across all tasks and time steps.
For instance, the action in Figure \ref{diagram:problem} would be \textit{type(ref=5, "username@email.com")}, where 5 refers to the index of the input when counted from top-to-bottom.
As model input, we concatenate the natural language instruction and \htmlpage\ into a single text input sequence.
Similarly, we treat the action as a text sequence for the model to predict.

\textbf{\taskClassification.} We use a dataset of 28K labelled examples, containing 66 different categories, of the form \textbf{(\htmlpage, element, category)}, previously used in the context of environment generation \citep{gur2021environment}. 
The dataset consists of HTMLs from real-world shopping websites and categories relevant to form-filling during payment and checkout on these websites.

\textbf{\taskDescription.} For this task, we derive a dataset from CommonCrawl.\footnote{\url{http://commoncrawl.org}} CommonCrawl does not include renderings or annotations that would reveal what text in the HTML is associated with which elements. Instead, we infer descriptions of various elements by exploiting a special attribute in the HTML schema known as \texttt{for}.
 As an example in Figure~\ref{diagram:problem}, the first \texttt{label} in the \htmlpage\ has a \texttt{for} attribute with value \textit{uName}, which
is the \texttt{id} of the element described by \texttt{label}; in this case, the \texttt{id} is that of the first \texttt{input} in the page.
This annotation does not affect the rendering of the page and is typically used for accessibility purposes. We utilize the information given by these \texttt{for} attributes to create a large-scale dataset to study description generation. A small sample is available in the supplemental material, while the entire dataset will be available upon publication.

Specifically, we collected 100 WARC (from April 2019) files from the CommonCrawl project and extracted all HTML \texttt{label}s that have a \texttt{for} attribute.
Removing non-Unicode and alphanumeric text in HTML \texttt{label}s results in a 400K example datset. We balance the distribution of labels, effectively downsampling the dataset to $85K$ samples. 
Each example is represented as \textbf{(\htmlpage, element, description)}, where \textbf{\htmlpage} is the HTML plaintext of the page, \textbf{element} is the element whose 
\texttt{id} attribute matches that appearing in the \texttt{label}'s \texttt{for} attribute, and \textbf{description} is the text inside the \texttt{label} element (see example in Figure~\ref{diagram:problem}). More details of the dataset can be found in Appendix \ref{sec:dataset_detail}.

\section{Pre-Processing}
\label{sec:preprocessing}
In treating HTML as token sequences, we minimize any HTML tree pre-processing prior to model input. We thus provide HTML as raw text (i.e., sequences of text tokens) and only apply a snippet extraction pre-processing for pages which are too large to fit into the typical \llms\ context windows.

\textbf{Snippet Extraction.} Real HTML pages can grow extremely large, reaching thousands of elements, far beyond the context window of the largest \llm\ that we studied (1920 tokens in PaLM \citep{chowdhery2022palm}).
\llms\ typically truncate such long sequences, which can be detrimental to HTML understanding as HTMLs are not linearly structured.
We take an element-centric approach and extract HTML snippets (a small portion of HTML code) surrounding a salient element (Figure~\ref{diagram:snippet_generation}).
A simple heuristic, which controls the tree's width and depth, guides the process: Start with a salient element and traverse its ancestors in the \htmlpage\ tree until a stopping condition is satisfied. As we traverse up, we estimate the height of the tree and the increased number of descendants of the new root.
We stop when either metric violates a pre-defined limit and take the resulting sub-tree as the snippet. 
We mark the salient element using a special attribute, called \textit{target}, to distinguish it from other elements.
We perform the snippet extraction for the semantic classification and description generation datasets, and keep the full HTML pages in MiniWoB because these pages are typically much smaller than real-world HTML.

\textbf{\htmlpage\ un-Parsing.}
We provide the models with the unparsed plaintext \htmlpage\ in the form of a sequence of tokens.
This canonical representation does not require specific model architectures such as hierarchical networks \citep{liu2018reinforcement,gur2021environment} and can be fed into any \llm.
We transform all datasets by converting every HTML page or snippet into a sequence.
For MiniWoB, we additionally concatenate (action history, instruction, \htmlpage) tuples into a single sequence.

\section{Model Training}
We study a variety of transformer-based \llms~\citep{vaswani2017attention} with different sizes and architectures for HTML understanding tasks (Table \ref{table:paper_summary}). In the rest of the text, we prefix models fine-tuned for \taskNavigation, \taskDescription, and \taskClassification\ with \modelN, \modelD, and \modelC,  respectively. For instance, \modelD-T5-3B is the three billion parameter T5 model \citep{raffel2020exploring} fine-tuned for the \taskDescription\ task. The rest of this section elaborates on training details. 

\textbf{Encoder-Decoder and Decoder-only Models.} 
We train encoder-decoder models, i.e.,  T5 \citep{raffel2020exploring}, and decoder-only models, i.e., LaMDA \citep{thooppilan2022lamda} and PaLM \citep{chowdhery2022palm}, with text input and text output (Figure~\ref{diagram:model}).
Inputs are raw HTML pages or snippet texts; similarly, outputs are categories, natural language descriptions, or actions represented as text.
Namely, for \taskClassification we use the textual representation of categories, similar to previous classification problems in NLP \citep{raffel2020exploring}.
For \taskNavigation, actions are converted into text by first converting them into key and value pairs and then concatenating the pairs.

Many websites in MiniWoB require multiple interactions, such as \textit{click-button-sequence} or \textit{click-checkboxes}, where each interaction might cause a subtle change in the website state.
For instance, after clicking on a checkbox in the \textit{click-checkboxes} website, its value flips from positive to negative or the other way around, which is not always reflected in \llms' predictions and leads to action repetitions.
We solve this issue by augmenting tuples in the dataset with a sequence of past actions, \textbf{(action history, instruction, HTML, action)}, and allowing \llms\ to learn from past experience.

\textbf{Encoder-only Models.} We train encoder-only models, i.e., BERT \citep{devlin2018bert}, with text input and categorical output.
We keep semantic categories as discrete one-hot classes.
To train encoder-only models, we add a new classification layer after the final encoder layer to produce a distribution over semantic categories.
In addition to the typical BERT models, we study MobileBERT \citep{sun2020mobilebert}, distilled from BERT-large with inverted bottlenecks, and Albert-XL \citep{lan2020ALBERT}, with parameter sharing and embedding split.

\section{Results}
\label{sec:results}
We now present the results of fine-tuned LLMs for HTML understanding.
We compare the models' performance with the existing baselines where possible (autonomous web navigation) and against other \llm\ architectures and training regimes (all tasks). Sections \ref{sec:resN}, \ref{sec:resC}, and \ref{sec:resD} evaluate task-specific performance, while Section \ref{sec:resAll} assesses the performance across all the tasks.   

\textbf{Metrics:} For autonomous web navigation we evaluate models' \textit{Success Rate}, which is averaged over 100 episodes per task.
For the other tasks, we use \textit{Accuracy} to measure exact match between prediction and ground truth. In the description generation task, we additionally provide evaluations using alternative `soft' text evaluation metrics, \textit{BLEU} and \textit{ROUGE-1}, measuring the similarity between predicted and ground truth text.

\begin{figure}
     \centering
     \begin{subfigure}[b]{0.45\textwidth}
         \centering
         \includegraphics[width=\textwidth]{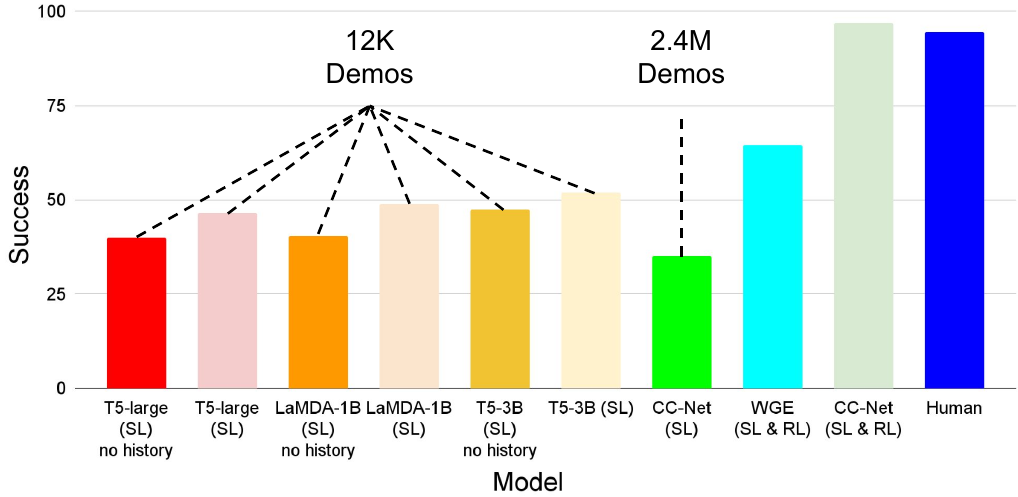}
         \caption{\small Baseline comparison.} 
         \label{fig:miniwob}
     \end{subfigure}
     \hfill 
     \begin{subfigure}[b]{0.45\textwidth}
         \centering
         {\small 
            \begin{tabular}{ l|c|cc|ccc }
             \textbf{Model Name}& \textbf{Success (\%)} & \textbf{Model Size}\\
             \hline
             \small {T5-large} & 18.1 & 800M \\
             \small {LaMDA-1B} & 15.6 & 1B\\
             \small {T5-3B} & 11.1 & 3B\\
             \hline
             \modelN T5-large & 46.4 & 800M\\
             \modelN LaMDA-1B & 48.8 & 1B\\
             \modelN T5-3B & 51.8 & 3B\\
            \end{tabular}
            }
         \caption{\small Pre-training effect.} 
         \label{table:transfer-webnav}
     \end{subfigure}
    
    \caption{\small a) \modelN-T5* performance compared to the previous SOTA models on MiniWoB benchmark. \modelN T5-3B improves the task success 16\% while using 192 times less data, compared to the best supervised learning (SL) model, CC-Net (SL). \llms\ performance is only surpassed by works utilizing RL, requiring orders of magnitude more online experience interaction with websites. b) \llms\ with and without pretraining on \taskNavigation\ task. Those with pretraining (denoted by the `\modelN' prefix) show a 2.5-4.5x performance improvement.}
    \label{fig:2}
\end{figure}

\subsection{Autonomous Web Navigation Results}
\label{sec:resN}
For \taskNavigation\ we fine-tune two \modelN\ encoder-decoder architectures (\modelN T5-large and \modelN T5-3B) on 12k demonstrations from human-annotated real websites. We evaluate the models on MiniWob \citep{liu2018reinforcement} benchmark, and compare with specialized architectures trained using supervised learning (SL) on 2.4 million human expert demonstrations \textit{CC-Net (SL)} \citep{humphreys2022data}, and two RL models bootstrapped with SL, CC-Net (SL) (CC-Net (SL \& RL) \citep{humphreys2022data}, and WGE (SL \& RL) \citep{liu2018reinforcement}). 
Additionally, we compare with the decoder-only architecture (\modelN Lambda-1B) and perform an ablation study on the impact of including the action history in the input.

\textbf{Comparison to SoTA.}
Since previous works report success on only a subset of websites in MiniWoB, we evaluate on 48 out of 62 websites that are common across all models. 
Table~\ref{table:miniwob-table} in the Appendix reports fine-grained results while Figure~\ref{fig:miniwob} presents results averaged over all websites.
Compared to CC-Net (SL) which is trained on all 2.4M demonstrations, \modelN T5-3B improves the success 16\% while only training on 12K publicly-available demonstrations, yielding over 192x improvement in sample-efficiency.
We find that all choices of \llms\ outperform previous SL models.
Notably, \modelN T5-3B significantly improves on websites requiring multiple-action sequences such as \textit{click\_checkboxes} or websites requiring entering text such as \textit{login\_user} (Table~\ref{table:miniwob-table}).
We observe that the performance of \llms\ is only surpassed by previous works utilizing RL, which require orders of magnitude more online experience interaction. Extending our fine-tuned \llms\ to an RL setting is a promising avenue for future work.

\textbf{Action history ablation.} 
Across all \llms\, we consistently observe a decrease in success, on average 6.4\%, when past actions are excluded from the inputs (Figure~\ref{fig:miniwob}).
Action history helps with websites that require entering multiple texts, as well as understanding minor changes that could be difficult to detect (e.g.
\textit{click\_checkboxes} and \textit{multi\_layout}).
\textit{multi\_layout} requires entering 3 different texts in the website where the layout is randomized at each episode, yet, surprisingly, even the (relatively smaller) \modelN T5-large model without action history outperforms the CC-Net (SL) model; illustrating that incorporating action history is not the only contributing factor for the better success.

\begin{table}
\centering
{\small
\begin{tabular}{ l|c|c|r | c }
 \textbf{Model Name}& \textbf{Test} (\%) & \textbf{Dev} (\%) & \textbf{Model Size} & \textbf{Code in training Corpus}\\
 \hline
 \modelC MobileBERT   & 78.1 & 77.7 & 24.6\,M & \multirow{10}{*}{0\%}\\
 \modelC Albert-XL   & 83.5 & 83.1 & 58.9\,M & \\
 \modelC BERT-smallest   & 84.4 & 83.6 & 38.7\,M & \\
 \modelC BERT-small   & 84.4 & 85.2 & 52.8\,M & \\
 \modelC BERT-medium   & 85.2 & 84.5 & 67\,M& \\
 \modelC BERT-base   & 83.9 & 84.8 & 109.5\,M&\\
 \modelC BERT-large   & 84.1 & 85.8 & 335.2\,M&\\
 \cline{1-4}
 \modelC T5-base   & 86.8 & 89.9 & 250\,M&\\
 \modelC T5-large   & 87.0 & 89.3 & 800\,M&\\
 \modelC T5-3B   & 87.7 & 90.3 & 3\,B&\\
 \hline
 \modelC LaMDA-1B & 87.4 & 87.1 & 1\,B & 12.5\% Code\\
 \modelC PaLM-8B   & 86.6 & 89.9 & 8\,B & 5\% Code (0.875\% HTML)\\
 \modelC PaLM-62B & \textbf{88.7} & \textbf{90.5} & 62\,B & 5\% Code (0.875\% HTML) \\
 \hline \hline
 T5-large & 76.4 & 75.2 & 800\,M  & \multirow{3}{*}{0\%}\\
 T5-3B  & 77.2 & 73.8 & 3\,B & \\
 PaLM-8B  & 73.3 & 70.1 & 8\,B & \\
\end{tabular}
}
\caption{\small \llms\ performance on the \taskClassification\ task. Fine-tuning off-the-shelf pretrained \llms\ (model names with prefix `Web*') helps \llms\ transfer better compared to training the same architecture from scratch on the HTML dataset (model names without prefix `Web*'), improving the accuracy of PaLM-8B more than 12\%. While \modelC PaLM-62B clearly performed better than all other models, we found \modelC T5-large to be competitive with much larger models such as \modelC LaMDA-1B or \modelC PaLM-8B.}
\label{table:classification_accuracy}
\end{table}
\subsection{Semantic Classification Task Results}
\label{sec:resC}
To evaluate the \taskClassification\ task, we compare the T5 encoder-decoder architecture's three size variants (\modelC T5-base, \modelC T5-large, and \modelC T5-3B) fine-tuned on 22K real, human-labeled training websites. We compare with a fine-tuned encoder only architectures (\modelC *BERT*), three fine-tuned decoder-only architectures (\modelC LaMDA and PaLM), and both encoder-decoder and decoder-only models trained on human labeled websites from scratch. Results are presented in Table-\ref{table:classification_accuracy}, where we find that all \modelC LLMs perform well and significantly better than the same architectures without pretraining.

\begin{figure}
\centering
\includegraphics[width=\textwidth]{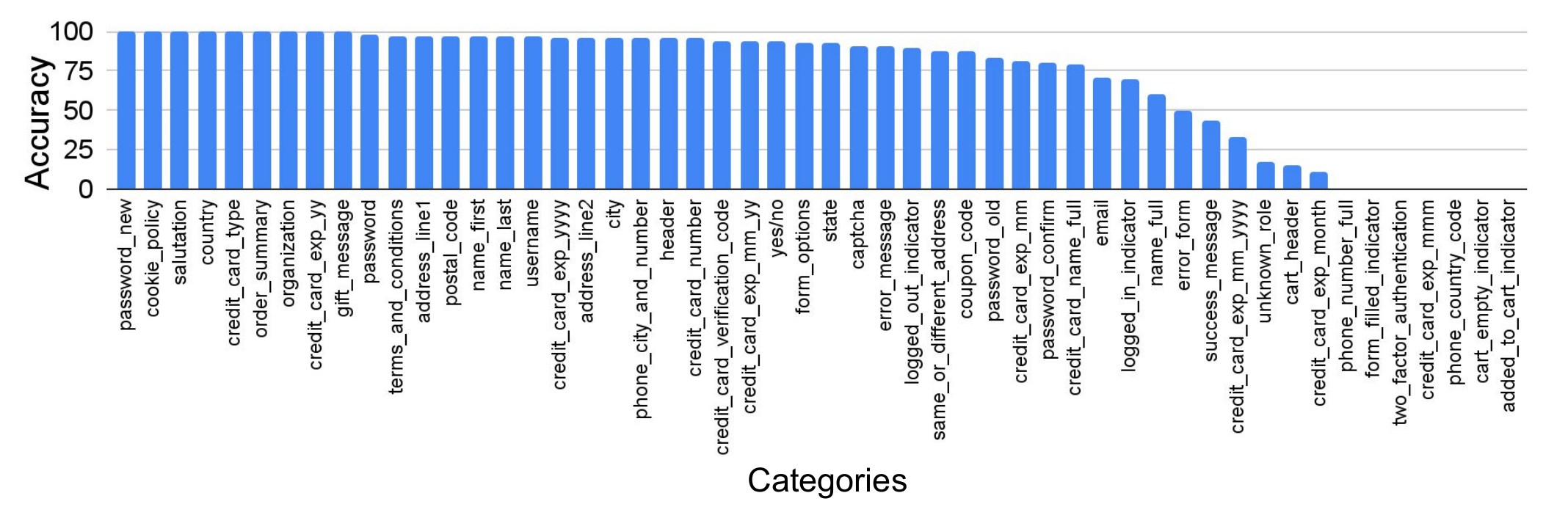}
\caption{\small Accuracy per classification category of the \modelC T5-3B model on the development dataset. }
\label{fig:acc-per-class}
\end{figure}

\begin{figure}
     \centering
     \begin{subfigure}[b]{0.45\textwidth}
         \centering
         {\small
            \begin{tabular}{ c|c|c|c }
             \textbf{{\scriptsize New}} &  \textbf{Height} & \textbf{Test} (\%)& \textbf{Dev} (\%)\\
             \textbf{{\scriptsize descendants}} (\%) & & &\\
              
             \hline
                25 & 3 & 87.7 & 90.3 \\
                25 & 4 & 88.6 & 89.2 \\
                50 & 3 & 88.4 & 90.0 \\
                50 & 4 & 89.3 & 89.2 \\
                \hline
                300 & 5 & 87.8 & 88.8 \\
                500 & 7 & 75.8 & 74.5 \\
            \end{tabular}
            }
         \caption{\small} 
         \label{table:ablation-snippet-gen}
     \end{subfigure}
     \hfill
          \begin{subfigure}[b]{0.45\textwidth}
         \centering
         \includegraphics[width=\textwidth]{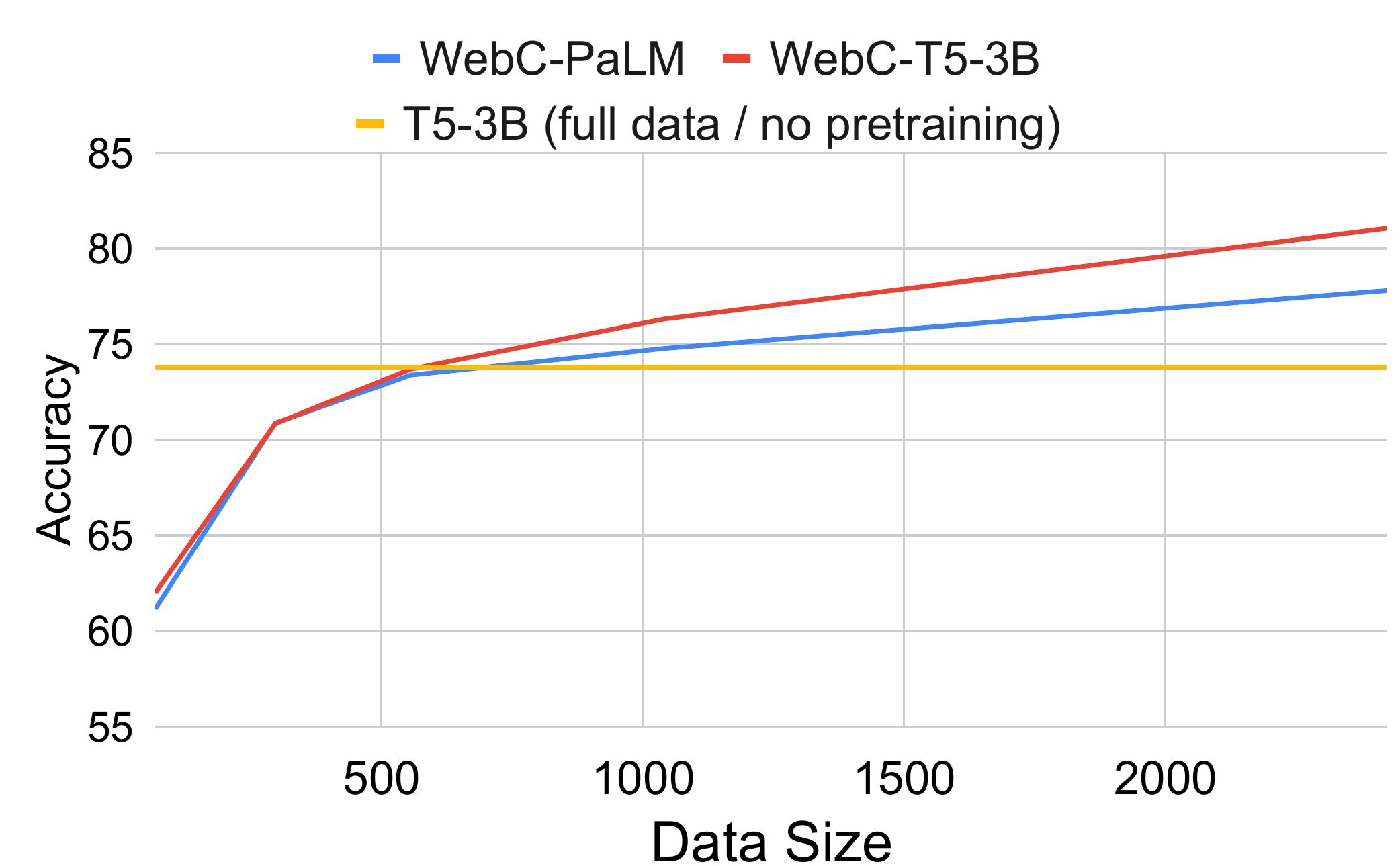}
         \caption{\small }
\label{fig:data-size-acc}
     \end{subfigure}
    \caption{\small a) Effect of snippet extraction parameters on \modelC T5-3B. Increases above 50\% in new descendants and height of 4. Large increases in both parameters lead to large snippets and decrease in accuracy. b) Accuracy over training data size. Using only 1000 labeled examples (4.4\% of all training dataset), \modelC T5-3B outperforms T5-3B (full data without pretraining) which is trained on \emph{all} available labeled data (approximately $30$k examples), and outperforms \modelC PaLM-8B which is an order of magnitude larger. }
    \label{fig:2}
\end{figure}

\textbf{Accuracy per category.}
In Figure~\ref{fig:acc-per-class}, we present accuracy distribution of the \modelC T5-3B model on the development dataset.
The fine-tuned encoder-decoder model performs strongly on a majority of the categories (Figure~\ref{fig:acc-per-class}), even on those with very few samples. For instance, the model is 100\% accurate on \textit{password\_new} which has only 56 training examples, because the class is unambiguous. On the other hand, unsurprisingly, the performance drops when the category is ambiguous, such as in the \textit{email} category which is frequently mistaken as \textit{username}.

\textbf{Snippet generation ablation.}
Two hyper-parameters govern snippet generation: percentage of new descendants and height of the new root. While small variations of both parameters do not change the performance, increasing both degrades the performance significantly (Table~\ref{table:ablation-snippet-gen}).
With new descendants up to 500\% and height up to 7, the performance drops by more than 15\%.
Note that snippet generation returns the full-page \htmlpage\ when both parameters increase indefinitely.

\textbf{Data size impact.}
When varying the fine-tuning training data sizes (1, 5, 10, 20, or 50 samples per class) in Figure~\ref{fig:data-size-acc}, 
\modelC T5-3B slightly outperforms \modelC PaLM-8B which is an order of magnitude larger.
Compared to T5-3B that is trained on all available HTML data without pretraining, \modelC T5-3B achieves better performance while using only 3.4\% of labeled data (1000 samples), thus highlighting the benefit of using standard off-the-shelf pretrained LLMs for HTML understanding.

\begin{table}[]
\centering
{\small 
\begin{tabular}{ l c|c|c|c|c|c  }
    & \multicolumn{3}{c}{\textbf{Test}} & \multicolumn{3}{|c}{\textbf{Dev}}\\
\cline{2-7} 
 \multicolumn{1}{l|}{\textbf{Model Name}}& \textbf{Accuracy}(\%) & \textbf{BLEU} & \textbf{ROUGE-1} & \textbf{Accuracy}(\%) & \textbf{BLEU} & \textbf{ROUGE-1}\\
 \hline
 \multicolumn{1}{l|}{\modelD T5-large}   & 83.2 & 90.2 & 90.5 & 84.3 & 91.7 & 91.5\\
\multicolumn{1}{l|}{\modelD  LaMDA-1B} & 83.3  & 87.5 & 90.2 & 84.3  & 88.6 & 91.2\\
 \multicolumn{1}{l|}{\modelD T5-3B}   &  84 & 90.8 & 90.9 & 85.2 & 92.1 & 91.9\\
 \hline
 \multicolumn{1}{l|}{Closest Description} & 57.4 & 24.4 & 59.2 & 60.8 & 23.9 & 62.1 \\

\end{tabular}
}
\caption{\small Description generation accuracy of \llms. }
\label{table:description-gen-acc}
\end{table}

\subsection{Description Generation Task Results}
\label{sec:resD}
For \taskDescription\, we split the CommonCrawl dataset based on URL top-level domains to test \llms' capabilities to generalize to unseen HTML.
We fine-tune encoder-decoder architectures (\modelD -T5*) and decoder-only models (\modelD -LaMDA*), with results presented in Table~\ref{table:description-gen-acc}. We also evaluate a strong heuristic baseline which simply finds the description closest to the salient element in the HTML text (Closest Description).

\textbf{Accuracy and Similarity Performance}
We show results of our evaluations in Table~\ref{table:description-gen-acc}.
All models achieve high scores across all metrics, achieving $\approx84\%$ on the accuracy in terms of exact match and a higher non-exact match score based on BLEU and ROUGE-1 ($\approx91\%$).
This difference indicates that the models are capable of locating the descriptions, but not always generating the exact output.

\subsection{HTML Understanding \llms\ Performance Analysis Across Tasks}
\label{sec:resAll}
We now analyze our results in aggregate to derive our main conclusions.
\subsubsection{Pretraining Effect: Pretraining on Large Text Corpora Matters}
\label{sec:respretrain}
Fine-tuned pretrained \llms\ outperform \llms\ trained on HTML-only data, improving the performance by more than 34.1\% on the \taskNavigation\ (Table~\ref{table:transfer-webnav}), and 10\% to 12.7\% on the \taskClassification\ task (Table~\ref{table:classification_accuracy}).

Since \taskNavigation\ is the most difficult task, the improved performance is an encouraging evidence of the value of \llms\ in HTML understanding tasks. Specifically, we observe that \llms\ without pretraining are comparable to fine-tuned pretrained models only on websites that require simple text matching.
In contrast, for websites such as \textit{click\_checkboxes}, text matching is harder and we find that pretraining is key to good performance.
We also found that without pretraining, model outputs were frequently in an incorrect format such as invalid dictionaries or invalid \text{ref}s with non-integer values. This suggests that the large corpora used for pretraining helps models to learn general HTML structure.

\subsubsection{Architecture Effect: T5-based Models Perform Best Across All Tasks} 
Encoder-decoder T5 based models perform better across all three tasks. On the \taskNavigation\ task, encoder-decoder (\modelN T5) architectures are better or comparable to \modelN LaMDA-1B (Figure~\ref{fig:miniwob}). On the \taskClassification, the smallest encoder-decoder model (\modelC T5-base) performs comparably to much larger decoder-only models (\modelC LaMDA-1B or \modelC PaLM-8B) and the largest encoder-only model (\modelC BERT-large) which has 85M more parameters (Table \ref{table:classification_accuracy}). We also observe that decoder-only PaLM-8B performs worse than much-smaller encoder-decoder T5-large when trained only on HTML data. Finally, on the \taskDescription\ encoder-decoder architecture has higher BLEU score.

One possible explanation for the strong performance of T5-based moels is the encoder-decoder architecture of these models. Namely, T5 models utilize an encoder with a bidirectional attention mechanism, not present in the LaMDA and PaLM decoders. The bidirectional attention mechanism can process HTML pages from both ends, potentially overcoming the loss of information when tree-structured HTML pages are converted into a fixed linear text sequences.

\subsubsection{Model Size Effect: Size (Sub-linearly) Matters}
\label{sec:ressize}
Across the tasks it appears that the architecture plays an important role in the model performance. Model size and performance are also positively correlated, although they reach diminishing returns. For instance, the model performance is roughly $O(\log \log n)$ with respect to model size on \taskClassification\ (Figure \ref{fig:data-size-acc} in Appendix). On the \taskNavigation\ task, performance grows slowly with the model size (Table~\ref{table:miniwob-table}), while on the \taskDescription\ it plateaus (Table~\ref{table:description-gen-acc}). 

\subsection{Discussion}
\label{sec:discussion}

\textbf{Bi-directional attention vs training corpora:} Pretraining on large corpora matters, yielding $\leq$4.5x performance improvements. Larger models tend to be better and we credit the bidirectional attention for T5's best overall performance across the tasks. PaLM and LaMDA include HTML and other code in their pretraining corpora, while BERT and T5 architectures did not, showing that pretraining on HTML is not necessary for strong performance when fine-tuned for HTML understanding. This strengthens the hypothesis behind the role of the bidirectional attention, and opens up the possibility to further improve the performance of T5 architectures by pretraining them on corpora with HTML. 

\textbf{Practical impact on labeling:} When available, the pretrained \llms\ need very little new expert data (200x and 30x reduction on the web navigation and classification tasks, respectively). This has a big potential impact on practical applications, reducing the data collection time and cost by orders of magnitude.

\textbf{Bigger is not always better:} When choosing the model size, the expected performance gains (sub-linear at best and asymptotic at worst) should be considered alongside the model's training and inference time and cost. For instance, on the classification task, the largest model \modelC PaLM-62B takes several days to fine-tune, and evaluates at 30\,Hz, while \modelC T5-large fine-tunes in several hours and evaluates at 700\,Hz -- an order of magnitude more expensive for a single percent uplift in accuracy. BERT models on the other hand train in minutes. If the application does not require high precision, these might be a good choice.

\textbf{Context window is a bottleneck:} The major bottleneck for the HTML understanding tasks seems to be the context window length that the current \llms\ support, even with models that accept 1000+ tokens.
It remains prohibitive to evaluate web navigation tasks on real websites that are orders of magnitude larger than pages in MiniWob. Similarly, we observed that increasing the snippet size leads to major performance degradation. This makes HTML understanding an interesting benchmark for future LLM development. For instance, new methods may need to be developed to compress the state representation of web content for use in LLM context windows.

\section{Conclusion}
\label{sec:conclusion}
We presented canonical tasks and fine-tuned \llms\ for HTML understanding. The comprehensive evaluations and analyses over a range of architectures, dataset sizes, and baselines yields practical findings and highlights current limitations of these models. We find that a) pretraining is critical for the performance and can reduce labeled data requirements, improving sample efficiency up to 200x; b) model architecture is the second-most important factor, and T5 models with bidirectional attention and encoder-decoder architecture perform the best across the board; c) given a choice, model size should be evaluated in the context of the model's training and inference performance, as the model size sub-linearly correlates with its performance. Finally, the proposed HTML understanding tasks highlight the relatively short context window that limits current LLMs, suggesting possibilities for future research that incorporate or eliminate this constraint.

\bibliography{iclr2022_conference}
\bibliographystyle{iclr2022_conference}

\appendix
\section{Appendix}
\subsection{Dataset Detail} \label{sec:dataset_detail}

Examining the description distribution, we found the original $400K$ dataset to be very skewed; only 20 descriptions (such as \textit{Email} and \textit{Password}) were covering 50\% of the dataset.
We sub-sampled the dataset so that each unique description has at most 10 data points.
We also found that \texttt{for} attributes are almost always defined for HTML \texttt{label}s. This could cause a model to overfit and just find the \texttt{label} element in the HTML and ignore everything else.
To avoid this sort of `cheating' we replace the tags of HTML \texttt{label}s by randomly sampling from \texttt{\{div, span, a, label\}}.
These tags are also frequently used to inject text in HTML but they are very rarely used with \texttt{for} attributes.
Finally, we removed examples where there are only a single text in the HTML since models can trivially generate descriptions by finding the only text in the HTML, which biases model weights and evaluation metrics.
After this final step, we have a total of $85K$ labeled examples.

\subsubsection{Snippet Generation}
\begin{figure}
\begin{center}
\includegraphics[width=\textwidth]{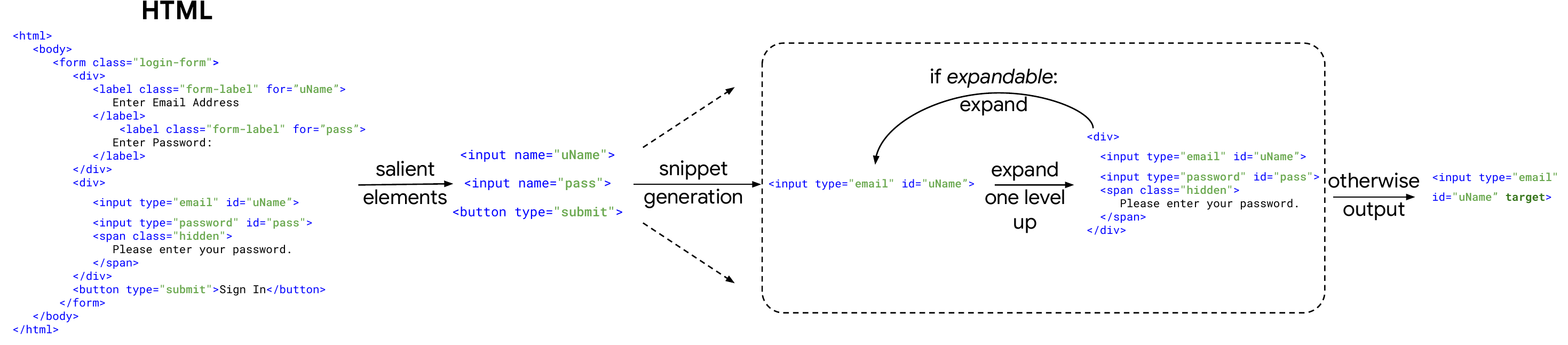}
\end{center}
\caption{\small High-level overview of our pre-processing pipeline for generating snippets from a full HTML webpage. Given the page, we detect salient elements and for each one of them we extract snippets by recursively moving up in the HTML tree until a validation heuristic fails. }
\label{diagram:snippet_generation}
\end{figure}

In Figure~\ref{diagram:snippet_generation}, we give a high-level overview of our snippet generation procedure.

\subsection{Additional Results}

\subsubsection{Semantic Classification}
\paragraph{Error Analysis.}
We manually examined 50 errors of T5-3B model over the development set (Table~\ref{table:error-type}) and assigned them into one of the 9 error types that we devised.
We found that 32\% of the errors are due to lack of information in the HTML snippets, which is mainly the result of lost information during snippet extraction process.
Annotation errors or email/username ambiguity make up 30\% of the errors.
These can't be improved without revising the annotated data or adding extra information to resolve the ambiguity.
We also found that the model sometimes picks a more general category, or a nearby text misleads the model; the latter usually happens when the HTML snippet is long where majority of the elements are noise.

\begin{table}[h]
\centering
\begin{tabular}{ |l|c| }
 \hline
 \textbf{Error Type}& \textbf{Percentage of Examples}\\
 \hline
 Not enough information in the HTML snippet & 30 \\
 \hline
 Incorrect annotation (ex: "unknown\_role" instead of "organization") & 12\\
 Annotation tool translates user selection incorrectly & 8 \\
 Email/Username ambiguity & 10\\
 \hline
 More general category (ex: "header" instead of "cart\_header") & 8 \\
 Immediate neighboring text misleads & 8 \\
 Incorrect date formatting (ex: "mm" instead of "mmm") & 4 \\
 No information in the HTML snippet & 2\\
 Others & 18 \\
 \hline
\end{tabular}
\caption{\small Types of errors over 50 manually examined examples. 32\% of errors are due to lack of information in HTML snippets, 30\% of errors are related to annotations or can't be improved due to ambiguity (email/username), and the remaining errors are incorrect predictions by the model.}
\label{table:error-type}
\end{table}

\begin{figure}[h]
\caption{Performance comparison w.r.t. increasing model size. As the model size increases, we observe an increase in overall accuracy with PaLM-62B model achieving the highest accuracy while being 7x larger than PaLM-8B.}
\centering
\includegraphics[width=0.7\textwidth]{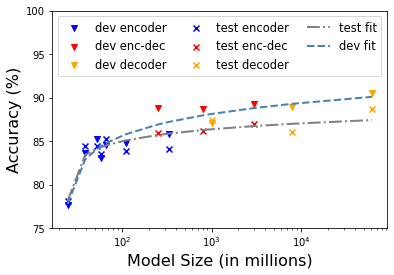}
\label{fig:model-size-acc}
\end{figure}

\textbf{Few-Shot Prompting}
In Table~\ref{table:fewshot-accuracy}, we present few-shot prompting performance of a 540B PaLM model.
We probe the model using a prompt template \texttt{<html> Role: <category>} with 1 example per category and generate categories using greedy-decoding.
In our preliminary experiments, we found that few-shot prompting achieves only 45.6 accuracy, much lower than a model fine-tuned on the same data (Figure~\ref{fig:model-size-acc}).
We found two common problems -- the model is not able to canonicalize predictions into categories and many of the examples are dropped due to context length.

\begin{wraptable}{h}{0.5\textwidth}
\centering
\begin{tabular}{ |l|c|c|c|  }
 \hline
 \textbf{Model Name}& \textbf{Test} & \textbf{Dev}\\
 \hline
 PaLM-540B & 64.2 & 60.3\\
 \hline
 - w/o Example Cleaning & 57.9 & 57.2\\
 \hline
 - w/o Category Rewriting & 52.1 & 50.7\\
 - w/o Dictionary Mapping & 45.6 & 45.1\\
 \hline
\end{tabular}
\caption{\small Few-shot prompting performance with different pre- and post-processing steps.  }
\label{table:fewshot-accuracy}
\end{wraptable}

We developed post-processing methods to alleviate the canonicalization problem and pre-processing methods to reduce lengths of examples.
Adding a dictionary-based mapping on predictions -- a manually curated paraphrase dictionary -- improves the performance to 52.1.
We also tried rewriting predictions by changing the order of tokens around "\_" such as \textit{name\_first} to \textit{first\_name} which further improved the performance to 57.9.
Finally, we cleaned examples in the prompt by removing certain elements such as \textit{"svg", "path", "img"}, and \textit{"iframe"} and also removing \textit{class} attribute from every element; this pre-processing step gives 64.2.

\subsection{Sample Episodes from MiniWoB}
See Table~\ref{tab:sample-ep} for an example episode of web navigation inferred by a fine-tuned LLM.

\begin{table}[h]
\centering
\caption{ A sample web page and corresponding episode using the T5-3B model. At each time step, previous actions, instruction, and \htmlpage\ are concatenated into a single \htmlpage\ text. Note that at the beginning of episode, there is no past actions and we simply concatenate instruction and \htmlpage. Action is generated as a sequence of tokens which is later parsed into a dictionary. The \textit{ref} in the action points to an element that has a \textit{ref} attribute with the same value. For instance, at the beginning of episode, \textit{ref: 6} corresponds to an input with \textit{ref=6}. At the end of the episode, the model clicks on the submit button and the episode terminates.}
\label{tab:sample-ep}
\begin{tabular}{cc}
\multicolumn{2}{c}{Web page} \\
\multicolumn{2}{c}{\includegraphics[width=0.2\textwidth]{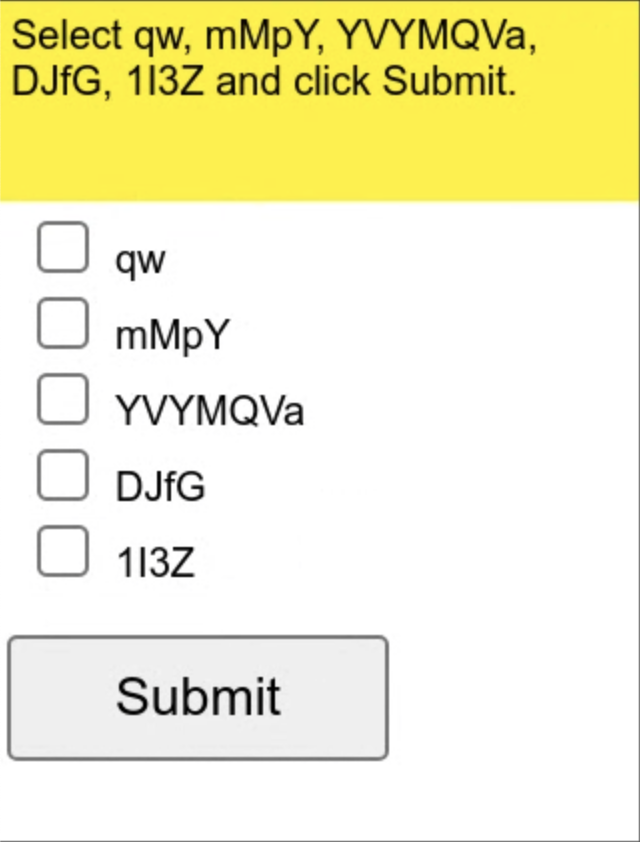}} \vspace{3mm} \\
\htmlpage\ Text & Action Text \\ \hline \hline 
 \includegraphics[width=0.7\textwidth]{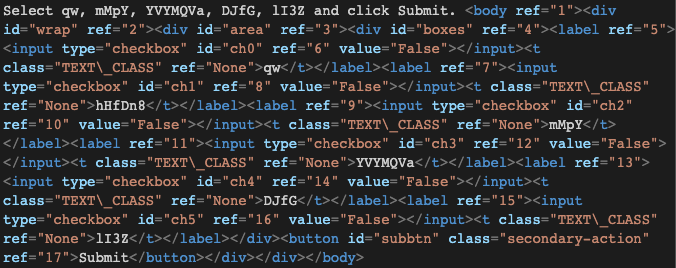} & \{action: click, ref: 6\} \\ \hline
 
\includegraphics[width=0.7\textwidth]{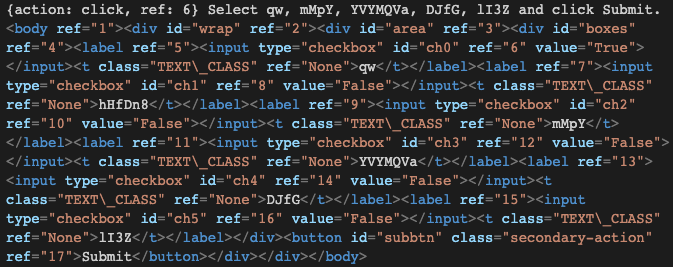} & \{action: click, ref: 10\}\\ \hline
 
\includegraphics[width=0.7\textwidth]{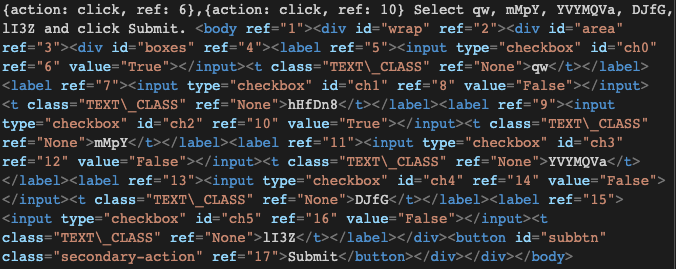} & \{action: click, ref: 12\}\\ \hline
  
\includegraphics[width=0.7\textwidth]{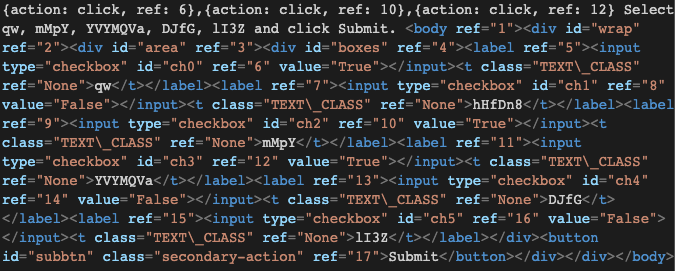} & \{action: click, ref: 14\}\\ \hline
\end{tabular}
\end{table}

\begin{table}[h]
\centering
\begin{tabular}{cc}

\includegraphics[width=0.7\textwidth]{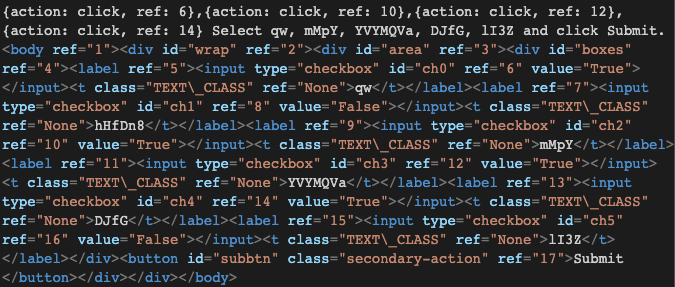} & \{action: click, ref: 16\}\\ \hline

\includegraphics[width=0.7\textwidth]{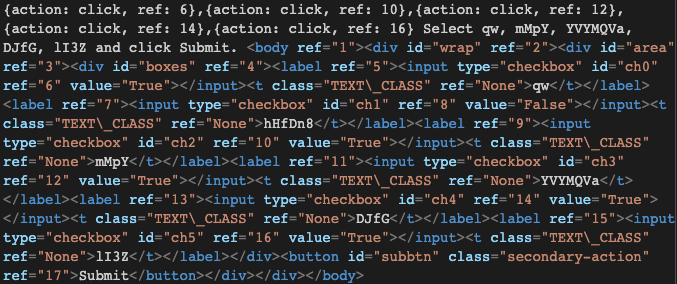} & \{action: click, ref: 17\}
\end{tabular}
\end{table}

\subsection{Detailed MiniWoB Results}
See Table~\ref{tab:miniwob-details} for detailed performance of various models on MiniWob.

\begin{table}[h]
\centering
\caption{\small Success rate comparison of various models in MiniWoB tasks. Baseline results are borrowed from \citep{humphreys2022data}. Note that these are normalized between 0 and 1. }
\label{tab:miniwob-details}
\resizebox{\textwidth}{!}{%
\begin{tabular}{|c|c|c|c|c|c|c|c|c|c|c|c|c|c|}
\hline
TASK & Human & & & CC-Net & CC-Net & World & Workflow & Learning & DOM-Q-Net & Workflow & Learning & Aggregated & Aggregated \\ 
 & & \modelN T5-3B & \modelN T5-3B  & (SL \& RL) & (SL) & of & guided & to & (RL) & guided & to & SOTA & SOTA \\
 & &  & (no history) & & & bits & exploration & navigate &  & exploration & navigate & (SL \& RL) & (Augmented) \\
 & & & & & & (SL \& RL) & (SL \& RL) & the web &  & (Augmented) & the web &  &  \\
 & & & & & & & & (RL) &  & & (Augmented) &  &  \\ \hline
bisect-angle & 0.92 & n/a & n/a & 0.97 & 0.29 & 0.8 & n/a & n/a & n/a & n/a & n/a & 0.8 & 0.8\\ \hline 
book-flight & 0.87 & 0 & 0 & 0.87 & 0 & 0 & 0 & n/a & n/a & 0 & 1 & 0 & 1\\ \hline 
chase-circle & 0.82 & n/a & n/a & 0.93 & 0.8 & 1 & n/a & n/a & n/a & n/a & n/a & 1 & 1\\ \hline 
choose-date-easy & 0.99 & 0.03 & 0.05 & 0.99 & 0.42 & n/a & n/a & n/a & n/a & n/a & n/a & n/a & n/a\\ \hline 
choose-date-medium & 0.98 & 0 & 0 & 0.99 & 0.26 & n/a & n/a & n/a & n/a & n/a & n/a & n/a & n/a\\ \hline 
choose-date & 0.97 & 0 & 0 & 0.97 & 0.12 & 0 & 0 & n/a & 1 & 0 & n/a & 1 & 1\\ \hline 
choose-list & 0.98 & 0.26 & 0.14 & 0.99 & 0.19 & 0.25 & 0.16 & 0.26 & n/a & 0.16 & 0.26 & 0.26 & 0.26\\ \hline 
circle-center & 0.96 & n/a & n/a & 0.97 & 0.36 & 0.98 & n/a & n/a & n/a & n/a & n/a & 0.98 & 0.98\\ \hline 
click-button-sequence & 0.94 & 1 & 1 & 1 & 0.47 & 0.22 & 0.99 & n/a & 1 & 1 & n/a & 1 & 1\\ \hline 
click-button & 0.98 & 1 & 0.96 & 1 & 0.78 & 0.62 & 1 & 1 & 1 & 1 & 1 & 1 & 1\\ \hline 
click-checkboxes-large & 0.87 & 0.22 & 0 & 0.71 & 0 & n/a & 0.68 & n/a & n/a & 0.84 & n/a & 0.68 & 0.84\\ \hline 
click-checkboxes-soft & 0.73 & 0.54 & 0.43 & 0.95 & 0.04 & n/a & 0.51 & n/a & n/a & 0.94 & n/a & 0.51 & 0.94\\ \hline 
click-checkboxes-transfer & 0.98 & 0.63 & 0.34 & 0.99 & 0.36 & n/a & 0.64 & n/a & n/a & 0.64 & n/a & 0.64 & 0.64\\ \hline 
click-checkboxes & 0.97 & 0.96 & 0.84 & 0.98 & 0.32 & 0.48 & 0.98 & n/a & 1 & 1 & n/a & 1 & 1\\ \hline 
click-collapsible-2 & 0.97 & 0 & 0.01 & 0.98 & 0.17 & 0.11 & 0.65 & n/a & n/a & 0.99 & n/a & 0.65 & 0.99\\ \hline 
click-collapsible & 0.99 & 0 & 0.01 & 1 & 0.81 & 0.98 & 1 & 1 & n/a & 1 & 1 & 1 & 1\\ \hline 
click-color & 0.97 & 0.27 & 0.23 & 1 & 0.82 & 0.23 & 1 & n/a & n/a & 1 & n/a & 1 & 1\\ \hline 
click-dialog-2 & 0.99 & 0.24 & 0.35 & 1 & 0.88 & 0.53 & 1 & n/a & n/a & 1 & n/a & 1 & 1\\ \hline 
click-dialog & 1 & 1 & 1 & 1 & 0.95 & 1 & 1 & 1 & 1 & 1 & 1 & 1 & 1\\ \hline 
click-link & 0.99 & 1 & 0.96 & 0.99 & 0.59 & 0.31 & 1 & 1 & 1 & 1 & 1 & 1 & 1\\ \hline 
click-menu-2 & 0.98 & n/a & n/a & 0.83 & 0.52 & 0.16 & n/a & n/a & n/a & n/a & n/a & 0.16 & 0.16\\ \hline 
click-menu & 0.97 & 0.37 & 0.38 & 0.94 & 0.22 & 0.13 & n/a & n/a & n/a & n/a & n/a & 0.13 & 0.13\\ \hline 
click-option & 0.99 & 0.87 & 0.78 & 0.99 & 0.21 & 0.28 & 1 & n/a & 1 & 1 & n/a & 1 & 1\\ \hline 
click-pie & 0.98 & 0.51 & 0.14 & 0.97 & 0.15 & 0.15 & 0.32 & 1 & n/a & 0.32 & 1 & 1 & 1\\ \hline 
click-scroll-list & 0.91 & 0 & 0 & 0.6 & 0.01 & 0.07 & n/a & n/a & n/a & n/a & n/a & 0.07 & 0.07\\ \hline 
click-shades & 0.91 & 0 & 0 & 1 & 0.04 & 0.27 & 0.22 & n/a & n/a & 0.99 & n/a & 0.27 & 0.99\\ \hline 
click-shape & 0.88 & 0.53 & 0.54 & 0.95 & 0.11 & 0.11 & 0.64 & n/a & n/a & 0.64 & n/a & 0.64 & 0.64\\ \hline 
click-tab-2-easy & 0.99 & n/a & n/a & 0.99 & 0.61 & n/a & n/a & n/a & n/a & n/a & n/a & n/a & n/a\\ \hline 
click-tab-2-hard & 0.96 & 0.12 & 0.13 & 0.98 & 0.19 & n/a & n/a & n/a & n/a & n/a & n/a & n/a & n/a\\ \hline 
click-tab-2-medium & 0.97 & n/a & n/a & 0.99 & 0.54 & n/a & n/a & n/a & n/a & n/a & n/a & n/a & n/a\\ \hline 
click-tab-2 & 0.97 & 0.18 & 0.09 & 0.98 & 0.27 & 0.08 & 0.64 & n/a & 1 & 0.98 & n/a & 1 & 1\\ \hline 
click-tab & 0.99 & 0.74 & 1 & 1 & 0.95 & 0.97 & 0.55 & 1 & 1 & 1 & 1 & 1 & 1\\ \hline 
click-test-2 & 0.99 & 1 & 1 & 1 & 0.95 & 0.83 & 1 & n/a & 1 & 1 & n/a & 1 & 1\\ \hline 
click-test-transfer & 0.99 & n/a & n/a & 1 & 0.94 & n/a & n/a & n/a & n/a & n/a & n/a & n/a & n/a\\ \hline 
click-test & 1 & 1 & 1 & 1 & 1 & 1 & 1 & n/a & 1 & 1 & n/a & 1 & 1\\ \hline 
click-widget & 0.83 & 1 & 0.97 & 1 & 0.56 & 0.34 & 0.93 & n/a & 1 & 0.93 & n/a & 1 & 1\\ \hline 
copy-paste-2 & 0.94 & n/a & n/a & 0.63 & 0.01 & 0 & n/a & n/a & n/a & n/a & n/a & 0 & 0\\ \hline 
copy-paste & 0.94 & n/a & n/a & 0.79 & 0.04 & 0 & n/a & n/a & n/a & n/a & n/a & 0 & 0\\ \hline 
count-shape & 0.82 & 0.41 & 0.43 & 0.85 & 0.21 & 0.18 & 0.59 & n/a & n/a & 0.76 & n/a & 0.59 & 0.76\\ \hline 
count-sides & 0.98 & n/a & n/a & 1 & 0.74 & 0.3 & n/a & n/a & n/a & n/a & n/a & 0.3 & 0.3\\ \hline 
drag-box & 0.99 & n/a & n/a & 1 & 0.61 & 0.31 & n/a & n/a & n/a & n/a & n/a & 0.31 & 0.31\\ \hline 
drag-cube & 0.99 & n/a & n/a & 0.79 & 0.23 & 0.18 & n/a & n/a & n/a & n/a & n/a & 0.18 & 0.18\\ \hline 
drag-item & 0.98 & n/a & n/a & 1 & 0.61 & n/a & n/a & n/a & n/a & n/a & n/a & n/a & n/a\\ \hline 
drag-items-grid & 0.87 & n/a & n/a & 0.98 & 0.05 & 0.01 & n/a & n/a & n/a & n/a & n/a & 0.01 & 0.01\\ \hline 
drag-items & 0.93 & n/a & n/a & 0.99 & 0.13 & 0.41 & n/a & n/a & n/a & n/a & n/a & 0.41 & 0.41\\ \hline 
drag-shapes & 0.96 & n/a & n/a & 0.99 & 0.26 & 0.92 & n/a & n/a & n/a & n/a & n/a & 0.92 & 0.92\\ \hline 
drag-sort-numbers & 0.92 & n/a & n/a & 0.97 & 0.11 & 0.66 & n/a & n/a & n/a & n/a & n/a & 0.66 & 0.66\\ \hline 
email-inbox-delete & 0.99 & n/a & n/a & 1 & 0.22 & n/a & n/a & n/a & 1 & n/a & n/a & 1 & 1\\ \hline 
email-inbox-forward-nl-turk & 0.88 & 0.33 & 0.09 & 1 & 0 & n/a & n/a & n/a & n/a & n/a & n/a & n/a & n/a\\ \hline 
email-inbox-forward-nl & 0.91 & 0.60 & 0.09 & 1 & 0 & n/a & n/a & n/a & n/a & n/a & n/a & n/a & n/a\\ \hline 
email-inbox-forward & 0.96 & n/a & n/a & 1 & 0.01 & n/a & n/a & n/a & n/a & n/a & n/a & n/a & n/a\\ \hline 
email-inbox-important & 0.99 & n/a & n/a & 1 & 0.3 & n/a & n/a & n/a & n/a & n/a & n/a & n/a & n/a\\ \hline 
email-inbox-nl-turk & 0.93 & 0.23 & 0.26 & 1 & 0.05 & n/a & 0.77 & n/a & n/a & 0.93 & n/a & 0.77 & 0.93\\ \hline 
email-inbox-noscroll & 0.96 & n/a & n/a & 1 & 0.13 & n/a & n/a & n/a & n/a & n/a & n/a & n/a & n/a\\ \hline 
email-inbox-reply & 0.91 & n/a & n/a & 1 & 0 & n/a & n/a & n/a & n/a & n/a & n/a & n/a & n/a\\ \hline 
email-inbox-star-reply & 0.95 & n/a & n/a & 1 & 0.11 & n/a & n/a & n/a & n/a & n/a & n/a & n/a & n/a\\ \hline 
email-inbox & 0.96 & 0.38 & 0.21 & 1 & 0.09 & 0.03 & 0.43 & n/a & 0.54 & 0.99 & n/a & 0.54 & 0.99\\ \hline 
enter-date & 0.97 & 0 & 0 & 1 & 0.02 & 0.61 & 0 & 1 & n/a & 0.96 & 1 & 1 & 1\\ \hline 
enter-password & 0.96 & 0.97 & 0.92 & 1 & 0.02 & 0 & 0.99 & 1 & 1 & 1 & 1 & 1 & 1\\ \hline 
enter-text-2 & 0.91 & n/a & n/a & 0.98 & 0.04 & 0 & n/a & n/a & n/a & n/a & n/a & 0 & 0\\ \hline 
enter-text-dynamic & 0.97 & 0.98 & 0.92 & 1 & 0.39 & 1 & 1 & 1 & 1 & 1 & 1 & 1 & 1\\ \hline 
enter-text & 0.98 & 0.89 & 0.99 & 1 & 0.35 & 0 & 1 & n/a & 1 & 1 & n/a & 1 & 1\\ \hline 
enter-time & 0.98 & 0 & 0.01 & 0.97 & 0.04 & 0.08 & 0.52 & n/a & n/a & 0.9 & n/a & 0.52 & 0.9\\ \hline 
find-midpoint & 0.94 & n/a & n/a & 0.97 & 0.35 & 0.31 & n/a & n/a & n/a & n/a & n/a & 0.31 & 0.31\\ \hline 
find-word & 0.96 & n/a & n/a & 0.88 & 0.05 & 0 & n/a & n/a & n/a & n/a & n/a & 0 & 0\\ \hline 
focus-text-2 & 0.99 & 1 & 1 & 1 & 0.96 & 0.83 & 1 & n/a & 1 & 1 & n/a & 1 & 1\\ \hline 
focus-text & 1 & 1 & 1 & 1 & 0.99 & 0.95 & 1 & n/a & 1 & 1 & n/a & 1 & 1\\ \hline 
grid-coordinate & 0.87 & 0.49 & 0.42 & 1 & 0.66 & 0.26 & 1 & n/a & n/a & 1 & n/a & 1 & 1\\ \hline 
guess-number & 0.99 & 0 & 0 & 1 & 0.21 & 0.2 & 0 & n/a & n/a & 0 & n/a & 0.2 & 0.2\\ \hline 
highlight-text-2 & 0.97 & n/a & n/a & 1 & 0.4 & 0.13 & n/a & n/a & n/a & n/a & n/a & 0.13 & 0.13\\ \hline 
highlight-text & 0.97 & n/a & n/a & 1 & 0.51 & 0.9 & n/a & n/a & n/a & n/a & n/a & 0.9 & 0.9\\ \hline 
identify-shape & 0.98 & 0.88 & 0.89 & 1 & 0.68 & 0.36 & 0.9 & n/a & n/a & 1 & n/a & 0.9 & 1\\ \hline 
login-user-popup & 0.94 & 0.72 & 0.40 & 1 & 0.02 & n/a & n/a & n/a & n/a & n/a & n/a & n/a & n/a\\ \hline 
login-user & 0.96 & 0.82 & 0.64 & 1 & 0 & 0 & 0.99 & 1 & 1 & 1 & 1 & 1 & 1\\ \hline 
moving-items & 0.18 & n/a & n/a & 0.88 & 0.13 & 0.78 & n/a & n/a & n/a & n/a & n/a & 0.78 & 0.78\\ \hline 
multi-layouts & 0.95 & 0.83 & 0.48 & 1 & 0 & n/a & 0.99 & n/a & n/a & 1 & n/a & 0.99 & 1\\ \hline 
multi-orderings & 0.96 & 0.88 & 0.64 & 1 & 0 & n/a & 0.05 & n/a & n/a & 1 & n/a & 0.05 & 1\\ \hline 
navigate-tree & 0.98 & 0.91 & 0.99 & 0.99 & 0.32 & 0.2 & 0.99 & 1 & 1 & 0.99 & 1 & 1 & 1\\ \hline 
number-checkboxes & 0.96 & n/a & n/a & 0.99 & 0 & 0.16 & n/a & n/a & n/a & n/a & n/a & 0.16 & 0.16\\ \hline 
read-table-2 & 0.95 & n/a & n/a & 0.94 & 0 & 0 & n/a & n/a & n/a & n/a & n/a & 0 & 0\\ \hline 
read-table & 0.97 & n/a & n/a & 0.97 & 0.01 & 0 & n/a & n/a & n/a & n/a & n/a & 0 & 0\\ \hline 
resize-textarea & 0.94 & n/a & n/a & 1 & 0.27 & 0.11 & n/a & n/a & n/a & n/a & n/a & 0.11 & 0.11\\ \hline 
right-angle & 0.87 & n/a & n/a & 0.98 & 0.26 & 0.38 & n/a & n/a & n/a & n/a & n/a & 0.38 & 0.38\\ \hline 
scroll-text-2 & 0.97 & n/a & n/a & 1 & 0.88 & 0.96 & n/a & n/a & n/a & n/a & n/a & 0.96 & 0.96\\ \hline 
scroll-text & 0.97 & n/a & n/a & 0.96 & 0.04 & 0 & n/a & n/a & n/a & n/a & n/a & 0 & 0\\ \hline 
search-engine & 0.97 & 0.34 & 0.34 & 1 & 0.15 & 0 & 0.26 & n/a & 1 & 0.99 & n/a & 1 & 1\\ \hline 
simon-says & 0.62 & n/a & n/a & 0 & 0.02 & 0.28 & n/a & n/a & n/a & n/a & n/a & 0.28 & 0.28\\ \hline 
simple-algebra & 0.86 & n/a & n/a & 0.75 & 0.03 & 0.04 & n/a & n/a & n/a & n/a & n/a & 0.04 & 0.04\\ \hline 
simple-arithmetic & 0.96 & n/a & n/a & 0.86 & 0.38 & 0.07 & n/a & n/a & n/a & n/a & n/a & 0.07 & 0.07\\ \hline 
social-media-all & 0.89 & 0 & 0 & 0.75 & 0 & n/a & 0.01 & n/a & n/a & 0.01 & 1 & 0.01 & 1\\ \hline 
social-media-some & 0.91 & 0.02 & 0 & 0.85 & 0.01 & n/a & 0.01 & n/a & n/a & 0.42 & n/a & 0.01 & 0.42\\ \hline 
social-media & 0.96 & 0.21 & 0.24 & 0.9 & 0.03 & 0.23 & 0.39 & n/a & 1 & 1 & n/a & 1 & 1\\ \hline 
terminal & 0.88 & n/a & n/a & -0.01 & 0 & 0 & n/a & n/a & n/a & n/a & n/a & 0 & 0\\ \hline 
text-editor & 0.88 & n/a & n/a & 0.98 & 0.11 & 0.01 & n/a & n/a & n/a & n/a & n/a & 0.01 & 0.01\\ \hline 
text-transform & 0.86 & n/a & n/a & 0.6 & 0.19 & 0 & n/a & n/a & n/a & n/a & n/a & 0 & 0\\ \hline 
tic-tac-toe & 0.71 & 0.48 & 0.40 & 0.83 & 0.32 & 0.34 & 0.37 & n/a & n/a & 0.47 & n/a & 0.37 & 0.47\\ \hline 
unicode-test & 0.99 & n/a & n/a & 1 & 0.86 & n/a & n/a & n/a & n/a & n/a & n/a & n/a & n/a\\ \hline 
use-autocomplete & 0.98 & 0.22 & 0.15 & 1 & 0.07 & 0 & 0.78 & n/a & n/a & 0.98 & n/a & 0.78 & 0.98\\ \hline 
use-colorwheel-2 & 0.94 & n/a & n/a & 0.95 & 0.38 & 1 & n/a & n/a & n/a & n/a & n/a & 1 & 1\\ \hline 
use-colorwheel & 0.9 & n/a & n/a & 0.98 & 0.68 & 1 & n/a & n/a & n/a & n/a & n/a & 1 & 1\\ \hline 
use-slider-2 & 0.97 & n/a & n/a & 0.95 & 0.03 & 0.15 & n/a & n/a & n/a & n/a & n/a & 0.15 & 0.15\\ \hline 
use-slider & 0.98 & n/a & n/a & 0.91 & 0.18 & 0.51 & n/a & n/a & n/a & n/a & n/a & 0.51 & 0.51\\ \hline 
use-spinner & 0.98 & 0.07 & 0.05 & 1 & 0.47 & 0.17 & 0.04 & n/a & n/a & 0.04 & n/a & 0.17 & 0.17\\ \hline 
visual-addition & 0.97 & n/a & n/a & 0.99 & 0.36 & 0.01 & n/a & n/a & n/a & n/a & n/a & 0.01 & 0.01 \\ \hline
\end{tabular}
}
\label{table:miniwob-table}
\end{table}

\subsection{Resource Requirements}
See Table~\ref{tab:resources}.

\begin{table}[h]
\centering
\caption{\small Resource requirements and running time of \llms. }
\label{tab:resources}
\resizebox{\textwidth}{!}{%
\begin{tabular}{c|c|c|c|c|c|c|c}
Model Name & Model Size & TPU version & Batch size & Input sequence length & Examples per sec (training) & Examples per sec (inference) \\
\hline
PaLM & 62B & TPU v4 & 8 & 1920 & 9.313 & 30.51 \\
PaLM & 8B & TPU v4 & 32 & 1920 & 64.4 & 184.3 \\ \hline
T5 & 3B & TPU v4 & 128 & 512 & 163.8 & 734.5 \\ \hline
LaMDA & 1B & TPU v2 & 128 & 512 & 363.1 & 1416 \\

\end{tabular}
}
\label{table:miniwob-table}
\end{table}

\subsection{Structure Dependence Ablation Study}

We conducted an ablation study to examine the sensitivity of model performance to preserving structural information. To do so, we evaluate the model’s performance on HTML input with critical structure components removed. We kept the order of elements and their attributes fixed while corrupting the nesting structure by removing closing tags. 

Removing closing tags corresponds to a valid traversal (BFS) and keeps the order of elements the same as the text based input.

As a simple example:

\begin{verbatim}
<div id=”form”><div><input id=”username”></div></div>
\end{verbatim}

would be converted into:

\begin{verbatim}
<div id=”form”><div><input id=”username”>
\end{verbatim}

We evaluated the trained WebN-T5-3B model on the same set of synthetic websites from the MiniWoB benchmark with this aspect of structure removed from the HTML pages. WebN-T5-3B achieves a 45.4\% success rate, 6\% lower than before, suggesting that WebN-T5-3B is at least partially dependent on the DOM topology. 

\subsection{Task-specific Models}

An alternative to LLMs is to adapt bespoke task-specific architectures tailored towards processing of structured documents and HTML (\cite{li2021markuplm, li2021structurallm}).

StructuralLM (\cite{li2021structurallm}) is an approach specifically tailored for document understanding (i.e., combinations of images and text), and thus makes several simplifying assumptions for its model that limit its applicability to HTML understanding (i.e., trees of elements with a richer structure and functionality). It is trained only on the textual content of a document - the markup information is ignored. For example, any input field or dropdown in a document would be missing from the model inputs. All of the tasks we study require knowledge of this information. For example, in autonomous navigation the model needs to interact with input elements (e.g. text, checkboxes, dropdowns) such as username and password in the login-user task in MiniWoB. Typically, a “type” action with a reference to an element and a text argument is generated by the model. Without knowing which input elements are available in the page, it is impossible to generate a reference to any input element. 

While MarkupLM (\cite{li2021markuplm}) is better tailored for understanding HTML pages, it has similar drawbacks as StructuralLM in that it focuses solely on text and structure of text while ignoring everything else in the markup. To illustrate our point better, we used the open source implementation of MarkupLM from the HuggingFace library (\cite{huggingface}) to process the sample HTML snippet in Figure-\ref{fig:1}(b). The MarkupLM ignores all input elements, both username and password, and generates \textit{\textless s\textgreater Email AddressEnter Password:Please enter your password.\textless /s\textgreater } which is the text input to the MarkupLM Transformer. Classifying this text as username or password is not possible without the additional context on which input element is the salient element (in this context it is the username). See below for the code to reproduce our result.

\begin{verbatim}
from transformers import MarkupLMProcessor
processor = MarkupLMProcessor.from_pretrained(f"microsoft/markuplm-base")
snippet = '''<div><label class="form-label" for=”uName”>Email Address
</label><label class="form-label" for=”pass”>Enter Password:
</label></div><div><input type="email" id="uName” target><input
type="password" id="pass"><span class="hidden">Please enter your password.
</span></div>'''
encoding = processor(snippet)
print(processor.batch_decode(encoding["input_ids"]))
\end{verbatim}

MarkupLM is also evaluated on NLP-like tasks such as QA or entity classification where understanding page content is paramount, whereas we focus on HTML understanding tasks such as autonomous navigation where both content and the page’s layout structure need to be understood. 

We perform a quantitative evaluation of MarkupLM on our tasks to understand how significant these limitations are. We fine-tune the MarkupLM-base model on the semantic classification task, using the same setup as other WebC models but with the suggested hyperparameters from (\cite{li2021markuplm}). We use the MarkupLM implementation from the HuggingFace library (\cite{huggingface}). On development and test sets, MarkupLM-base achieves 65\% and 66\% accuracy, respectively. These results are more than 16\% lower compared to similar size WebC-BERT-base results that we report in our work. This suggests that although domain specific models may be suitable for processing HTML for NLP tasks, the generality, flexibility, and sample efficiency LLMs provide advantages for autonomous navigation tasks.

\end{document}